\documentclass[sigconf]{acmart}
\AtBeginDocument{%
  \providecommand\BibTeX{{%
    \normalfont B\kern-0.5em{\scshape i\kern-0.25em b}\kern-0.8em\TeX}}}

\copyrightyear{2021}
\acmYear{2021}
\setcopyright{iw3c2w3}
\acmConference[WWW'21]{Proceedings of the Web Conference 2021}{April 19--23, 2021}{Ljubljana, Slovenia}
\acmBooktitle{Proceedings of the Web Conference 2021 (WWW'21), April 19--23, 2021, Ljubljana, Slovenia}
\acmPrice{}
\acmDOI{10.1145/3442381.3449898}
\acmISBN{978-1-4503-8312-7/21/04}

\usepackage{amsmath} %
\usepackage{algorithm} %
\usepackage{algorithmic} %
\usepackage{epstopdf}
\usepackage{appendix} %
\usepackage{multirow} %
\usepackage{graphicx} %
\usepackage{enumitem}
\usepackage{bm}  %
\usepackage{soul,color} %
\soulregister\cite7
\soulregister\ref7
\soulregister\pageref7
\newcommand{\tabincell}[2]{\begin{tabular}{@{}#1@{}}#2\end{tabular}}

\begin{document}

\title{MulDE: Multi-teacher Knowledge Distillation for Low-dimensional Knowledge Graph Embeddings}

\author{Kai Wang}
\authornote{Corresponding author}
\affiliation{%
  \institution{School of Software, the Key Laboratory for Ubiquitous Network and Service Software of Liaoning Province, Dalian University of Technology}
  \city{Dalian}
  \state{Liaoning}
  \country{China}
  \postcode{116620}}
\email{kai_wang@mail.dlut.edu.cn}

\author{Yu Liu}
\affiliation{%
  \institution{School of Software, the Key Laboratory for Ubiquitous Network and Service Software of Liaoning Province, Dalian University of Technology}
  \city{Dalian}
  \state{Liaoning}
  \country{China}
  \postcode{116620}}
\email{yuliu@dlut.edu.cn}

\author{Qian Ma}
\affiliation{%
  \institution{School of Software, the Key Laboratory for Ubiquitous Network and Service Software of Liaoning Province, Dalian University of Technology}
  \city{Dalian}
  \state{Liaoning}
  \country{China}
  \postcode{116620}}
\email{Ninja666@mail.dlut.edu.cn}

\author{Quan Z. Sheng}
\affiliation{%
 \institution{Intelligent Computing Laboratory\\ Department of Computing\\ Macquarie University}
 \city{Sydney}
 \state{NSW}
 \country{Australia}
 \postcode{2109}}
 \email{michael.sheng@mq.edu.au}
 
\renewcommand{\shortauthors}{Wang et al.}

\begin{abstract}
Link prediction based on knowledge graph embeddings (KGE) aims to predict new triples to automatically construct knowledge graphs (KGs).
However, recent KGE models achieve performance improvements by excessively increasing the embedding dimensions, which may cause enormous training costs and require more storage space.
In this paper, instead of training high-dimensional models, we propose MulDE, a novel knowledge distillation framework, which includes multiple low-dimensional hyperbolic KGE models as teachers and two student components, namely Junior and Senior. 
Under a novel iterative distillation strategy, the Junior component, a low-dimensional KGE model, asks teachers actively based on its preliminary prediction results, and the Senior component integrates teachers' knowledge adaptively to train the Junior component based on two mechanisms: {\em relation-specific scaling} and {\em contrast attention}. 
The experimental results show that MulDE can effectively improve the performance and training speed of low-dimensional KGE models. The distilled 32-dimensional model is competitive compared to the state-of-the-art high-dimensional methods on several 
widely-used datasets.
\end{abstract}



\maketitle

\section{Introduction}
Knowledge graphs (KGs), which describe human knowledge as factual triples in the form of (head entity, relation, tail entity), have shown great potential in various domains \cite{IJCAI19MKG, WWW20KG3, WWW20KGAdd2}. 
Popular real-world KGs such as Yago \cite{Yago} and DBPedia \cite{DBpedia}, which typically contain an enormous quantity of entities, are still far from complete \cite{TransE, WWW20KG2}.
Due to the high manual costs of discovering new triples \cite{TripleCost}, link prediction based on knowledge graph embeddings (KGE) \cite{AAAI20KG,WWW20KG} has drawn considerable attention very recently 
as a means to overcome this problem.
A typical KGE model first represents entities and relations as trainable continuous vectors. Then, given an entity and a relation of a triple (we call `e-r query'), the model defines a scoring function to measure each candidate in the entity set and outputs the best one  \cite{2017Survey}.

To achieve higher prediction accuracy, recent KGE models generally use high-dimensional embedding vectors up to 200 or even 500 dimensions \cite{RotatE, QuatE}. 
However, when we have millions or billions of entities in a KG, the high-dimensional model 
demands enormous training costs and storage space \cite{KGCompress, WWW20KGAdd1}. 
This prevents downstream AI applications from updating KG embeddings promptly or being able to be deployed on mobile devices. 
Recently, several research efforts have drawn attention to this research issue by either improving the low-dimensional models (such as 8 or 32 dimensions) \cite{GoogleAttH}, or compressing pre-trained high-dimensional models \cite{KGCompress}. However, the former cannot utilize the high-accuracy knowledge from high-dimensional models, and the latter suffers from high pre-training costs and cannot continue training when the KG is modified.

Can we transfer high-accuracy knowledge to a low-dimensional model while avoiding to train high-dimensional models? 
According to a preliminary experiment, we find that the ensemble of different 64-dimensional hyperbolic KGE models can outperform any single higher-dimensional models.
To this end, we determine to employ those low-dimensional hyperbolic models as multiple teachers and integrate their knowledge to train a smaller KGE model in a Knowledge Distillation (KD) process. 
Knowledge Distillation \citep{1stKD}, which has been rarely applied in the knowledge graph domain, is a technology distilling `soft labels' from a pre-trained big model (teacher) to train a small one (student). In our paper, soft labels come from the scores of candidate triples measured by the KGE model.
Compared with a single high-dimensional teacher, we argue that there are at least three benefits of utilizing multiple low-dimensional teachers:
\begin{itemize}
\item {\em Reduce pre-training costs}. The number of parameters for multiple teachers is relatively lower than that of a high-dimensional state-of-the-art model. Besides, the pre-training speed of the former can be further improved by parallel techniques.
\item {\em Guarantee teacher performance}. The ensemble's accuracy exceeds some high-dimensional models, because prediction errors of a single teacher can be corrected by integrating multiple prediction results.
\item {\em Improve distilling effect}. Based on recent knowledge distillation studies \cite{AAAI20KD}, a student model is easier to acquire knowledge from a teacher with a similar size. A teacher in low dimensions is more suitable than the one having hundreds of dimensions.
\end{itemize}

In this paper, we pre-train multiple low-dimensional hyperbolic KGE models, and propose a novel framework utilizing \textbf{Mul}ti-teacher knowledge \textbf{D}istillation for knowledge graph \textbf{E}mbeddings, named \textbf{MulDE}. 
Different from conventional KD technologies, we design a novel {\em iterative distillation strategy} and two student components (i.e., {\em Junior} and {\em Senior}) in MulDE. 
In one iteration, the Junior component first predicts top-K candidates for each e-r query, and transfers to multiple teachers. 
Then, the Senior component integrates teacher results and generates soft labels with two mechanisms: {\em relation-specific scaling} and {\em contrast attention}. 
Finally, Junior receives soft labels and updates its parameters using a knowledge distillation loss. 
As a result, instead of receiving knowledge in one direction from teachers, the students can seek knowledge actively from teachers and thereby distinguish false-positive entities better.

We conduct extensive experiments on two widely-used datasets, FB15k-237 and WN18RR to validate our proposed model. The results show that MulDE can significantly improve the prediction accuracy and training speed of low-dimensional KGE models. 
The distilled 32-dimensional model outperforms the state-of-the-art low-dimensional models, and is comparable to some high-dimensional ones. 
According to ablation experiments, we prove the effectiveness of the iterative distillation strategy and the other major modules in MulDE. 
We also compare different teacher settings and conclude that using four different 64-dimensional hyperbolic models is optimal.

The rest of the paper is organized as follows. 
We briefly introduce the background and notations in our work 
in Section \ref{sec:2}. Section \ref{sec:3} details the whole framework and basic components of the MulDE model. 
Section \ref{sec:4} reports the experimental studies, and Section \ref{sec:5} further discusses the experimental investigations. We 
discuss the related work in Section \ref{sec:6} and, finally, offer some concluding remarks in Section \ref{sec:7}.

\section{Preliminaries}
\label{sec:2}

\subsection{Definitions and Notations}
Let $E$ and $R$ denote the set of entities and relations,
a knowledge graph $\mathcal{G}$ is a collection of factual triples $(e_h,r,e_t),$ where $e_h,e_t \in E$ and $r \in R$. $N_e$ and $N_r$ refer to the number of entities and relations, respectively.
Given an e-r query $q = (e_{in}, r)$, where $e_{in} \in E$ and $r \in R$, the link prediction task is to find $e_{miss} \in E$, such that $(e_{in}, r, e_{miss})$ or $(e_{miss}, r, e_{in})$ should belong to the knowledge graph $\mathcal{G}$.
According to the Cartesian product, the number of all possible e-r queries is equal to $N_e \times N_r$. 

Knowledge Graph Embeddings aim to represent each entity $e \in E$ and each relation $r \in R$ as $d$-dimensional continuous vectors.
A KGE model $\mathcal{M}$ defines a scoring function $F : E \times R \times E \rightarrow \mathcal{R}$ to score each triple through embedding vectors. Given an e-r query $q$, $\mathcal{M}$ outputs a prediction sequence containing sorted scores of all candidate triples, denoted as $\{(c,s)|c \in E, s \in \mathcal{R}\}$. In the sequence, candidates with higher scores are more likely to be true.

For each candidate triple $(e_{in},r,e_{c})$, $\mathcal{M}$ learns the relation vector $\bm{r}$ as a transformation between two entity vectors $\bm{e_{in}}$ and $\bm{e_c}$.
From a probabilistic model point of view, regarding the e-r query vector $\bm{q}$ as a transformed entity vector, the plausibility of the triple can be expressed as a joint probability $p(q, e_{c})$:
\begin{align}
\label{eq1}
   p\left(q, e_c\right)=\frac{e^{D(\bm{q},\bm{e_c})}}{\sum_{i=1}^{N_e \times N_r} \sum_{j=1}^{N_e} e^{D( \bm{q_i},\bm{e_j} )}}
\end{align}
where $\bm{e} \in \mathcal{R}^d$ is the embedding vector of an entity $e$, the query vector $\bm{q} = \bm{e_{in}} \circ \bm{r}$, $\circ$ refers to any vector transformation, and $D(\bm{q}, \bm{e})$ refers to the distance between two entity vectors in the scoring function $F$.

\begin{figure}[!bt]
\centering
\includegraphics[width=0.35\textwidth]{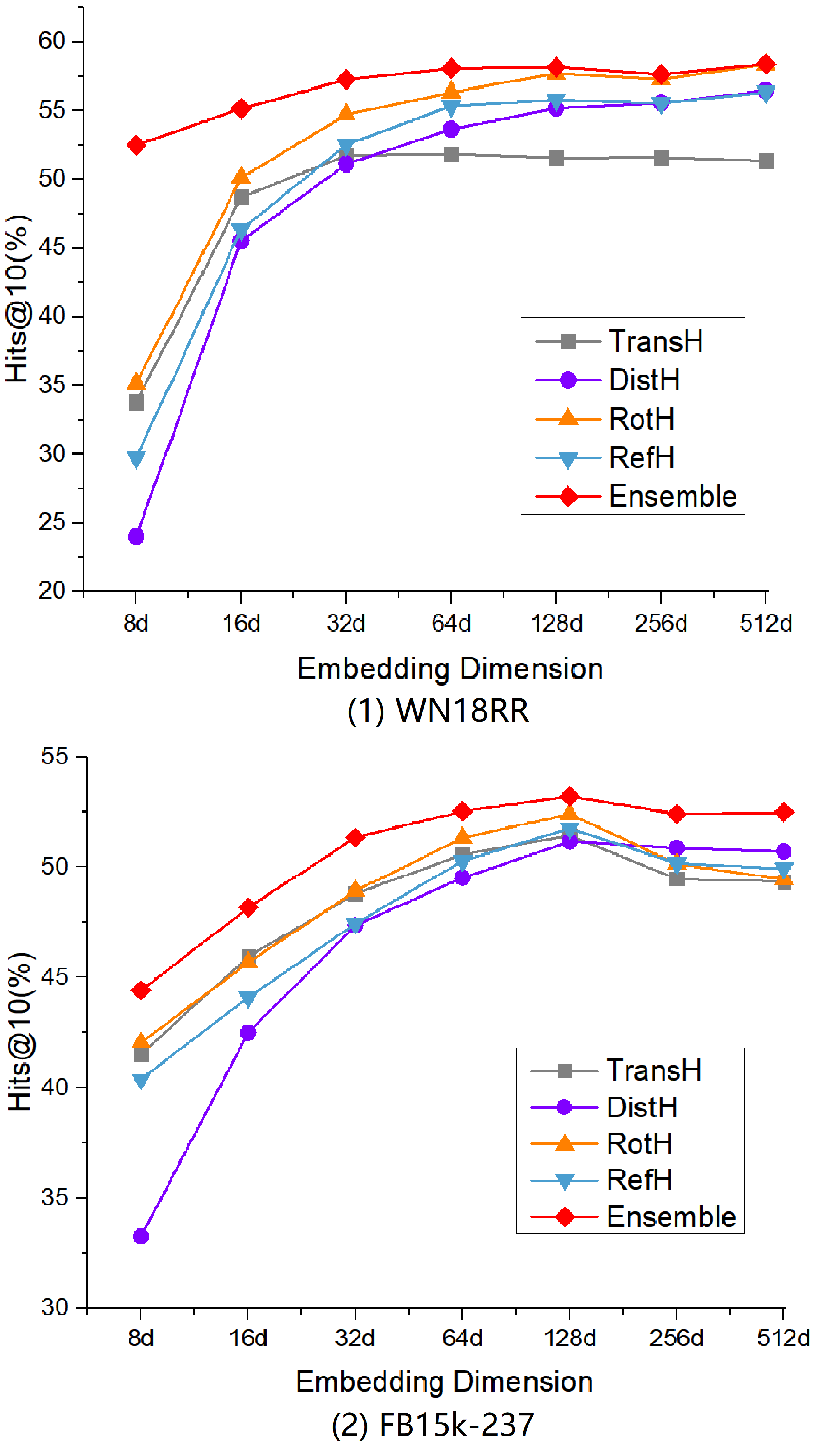}
\caption{The changes of model performance (Hits@10) with the growth of the embedding dimensions on two different datasets. The Ensemble model adds the prediction results of the four models directly.}
\label{fig:1}
\vspace{-5mm}
\end{figure}

\subsection{Preliminary Analysis of Low-Dimensional Embeddings}
In a KGE model, the parameters of embedding vectors usually account for most of the total parameters. 
Given a large-scale KG, reducing the embedding dimensions can significantly save storage space.
However, small embedding dimensions lead to low representation capability, limiting the prediction accuracy of the KGE model.
The information quantity of a low-dimensional embedding vector space is not infinite, and cannot accommodate the information of numerous triples when facing millions of entities.

To the best of our knowledge, there is no theoretical analysis of the minimum embedding dimensions required by a knowledge graph.
From the view of the information entropy, 
a KG $\mathcal{G}$ has a certain degree of uncertainty. An e-r query points to one or more target entities. Meanwhile, given a large-scale entity set, there will be hidden triples that are not discovered inevitably. On this basis, when using a KGE model $\mathcal{M}$ to encode $\mathcal{G}$, we can train $\mathcal{M}$ to minimize its information entropy, thereby ensure $\mathcal{M}$ is sufficient to keep the information of the KG. 

According to the above analysis, we argue that there is a sufficient condition to ensure $\mathcal{M}$'s entropy $H_\mathcal{M}$ is smaller than the entropy of KG $H_\mathcal{G}$. However, the information entropy of both sides is hard to measure.
Different knowledge graphs contain various entity types and topological structure, while a KGE model has its specific scoring function and randomly initialized parameters.
These factors hinder measuring the minimum dimension of a knowledge graph by theoretical derivation.
To this end, we conduct a preliminary experiment to analyze the possible value of the minimum dimension on two specific KG datasets, WN18RR \cite{WNDS} and FB15k237 \cite{Toutanova2015}.
Focusing on the low-dimensional KGE task, we employ four different hyperbolic KGE models \cite{GoogleAttH} and compare their performance with different embedding dimensions. The experimental results are shown in Fig. \ref{fig:1}. 

There are two inspirations we find from this experiment:
(1) On the two datasets, the prediction accuracy of all four models keeps increasing with the growth of embedding dimensions until around 64.
When the embedding dimensions are more than 64, the model performance only slightly improves or starts to fluctuate. 
This indicates that 64-dimensional models can already achieve high performance while using much smaller parameters than those high-dimensional ones.
(2) When integrating the scores of all hyperbolic models, the 64-dimensional ensemble is already better than any higher-dimensional models.
Therefore, instead of training a high-dimensional model and compressing,
we can utilize the low-dimensional ensemble as a lightweight source of high-accuracy knowledge.

Overall, this preliminary analysis indicates that 
a 
64-dimensional space has an excellent capability to represent a normal-size KG dataset,
which provides a guidance for the selection of teacher dimensions in our framework.
Furthermore, inspired by the ensemble's results, we focus on utilizing multiple low-dimensional models as teachers to reduce training costs while ensuring prediction accuracy.


\section{Methodology}  
\label{sec:3}   

\begin{figure}[!bt]
\centering
\includegraphics[width=1\linewidth]{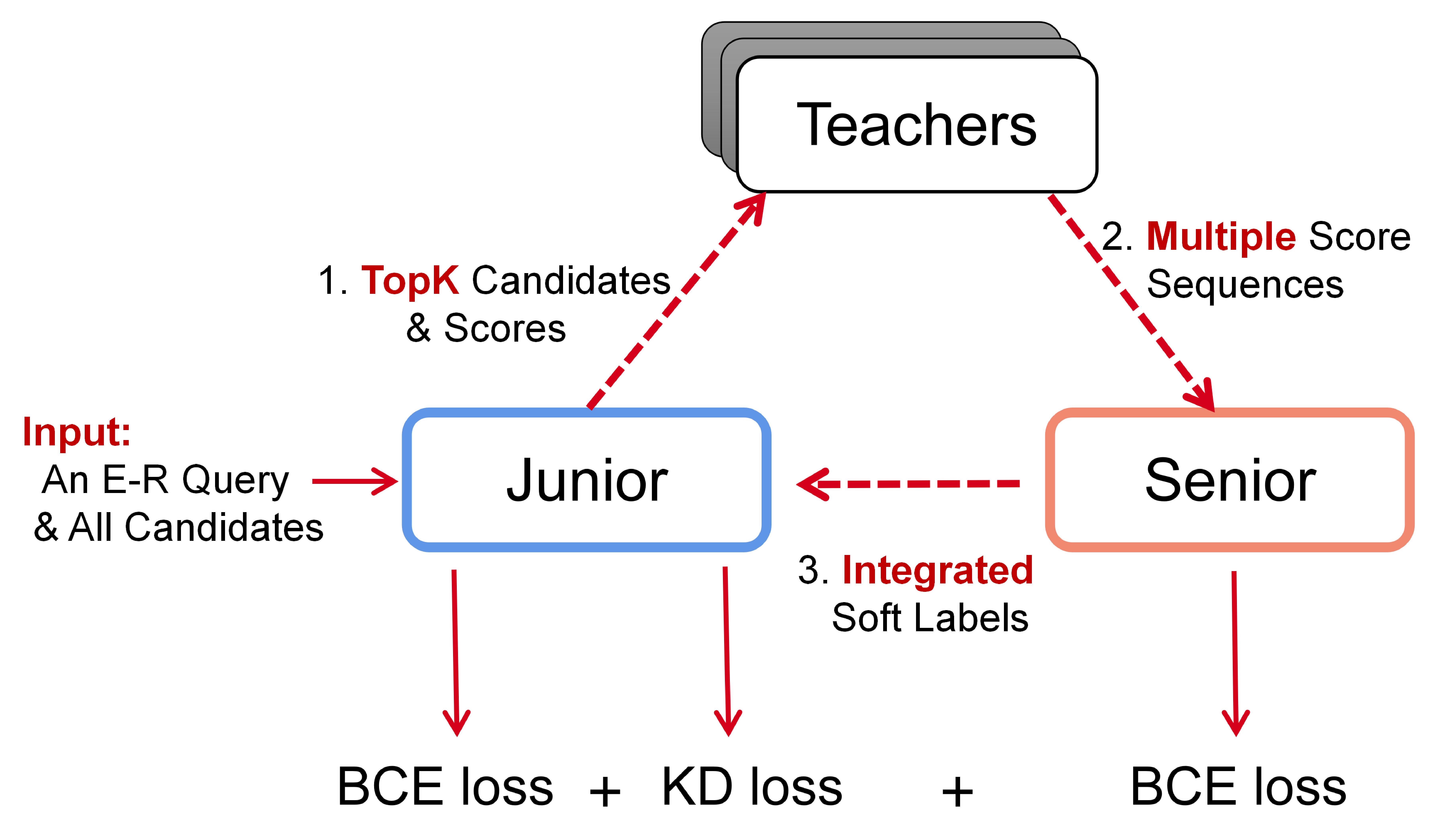}
\caption{An illustration of the MulDE framework. `BCE loss' refers to the binary cross-entropy loss, and `KD loss' refers to the loss function of knowledge distillation.}
\label{fig:2}
\end{figure}

The proposed Multi-teacher Distillation Embedding (MulDE) framework utilizes multiple pre-trained low-dimensional models as teachers. Under a novel iterative distillation strategy, it integrates prediction sequences from different teachers, and supervises the training process of a low-dimensional student model.
Figure \ref{fig:2} illustrates the architecture of the MulDE framework, including multiple teachers and two student components (Junior and Senior):

\begin{itemize}
\item{\bf{Multiple teachers}} are regarded as the data source of prediction sequences, and have no parameter updates in the training process. We employ four different hyperbolic KGE models as teachers. The details about pre-trained teachers will be described in Section \ref{sec:3.2}. 
\item{\bf{The Junior component}} is the target low-dimensional KGE model.
Given an e-r pair, the Junior component sends its top-K predicted entities to teachers, and gets the corresponding soft labels from the Senior component.
We will detail the Junior component in Section \ref{sec:3.3}.
\item{\bf{The Senior component}} acquires prediction sequences from teachers, and then generates soft labels through two mechanisms: {\em relation-specific mechanism} and {\em contrast attention mechanism}.
The details of the Senior component will be discussed in Section \ref{sec:3.4}.
\end{itemize}

As shown in Fig. \ref{fig:2}, unlike traditional one-way guidance, the iterative distilling strategy in MulDE forms a novel {\em circular interaction} between students and teachers. 
In each iteration, the Junior component makes a preliminary prediction based on an e-r query, and selects those indistinguishable entities (top-K) to ask multiple teachers. In this way, Junior can effectively correct its prediction results and outperform the other models in the same dimension level. Meanwhile, rather than processing the fixed teacher scores, the Senior component can adjust parameters continually according to the Junior's feedback, and generate soft labels according to training epochs and the Junior's performance adaptively. The learning procedure of MulDE will be detailed in Section \ref{sec:3.5}.

\subsection{Pre-trained Teacher Models}
\label{sec:3.2}   
To ensure the performance of the low-dimensional teachers, we employ a group of hyperbolic KGE models proposed by Chami et al. \cite{GoogleAttH}, including TransH, RefH, and RotH. 

Taking the RotH model as an example, it uses a $d-$dimensional Poincar\'{e} ball model with trainable negative curvature. Embedding vectors are first mapped into this hyperbolic space, and a relation vector is regarded as a rotation transformation of entity vectors.
Then, RotH employs a hyperbolic distance function to measure the difference of transformed head vector and candidate tail vector. 
Similarly, TransH uses M\"{o}bius addition to imitate TransE \cite{TransE} in hyperbolic space, and RefH replaces the rotation transformation with a reflection one.

Considering the effectiveness of aforementioned models using inner product transformation, such as DistMult \cite{DistMult} and ComplEx \cite{ComplEx}, we add another hyperbolic model named DistH into this group to get more accurate prediction sequences. The scoring functions of these models are as follows:
\begin{align}
   &TransH: F(e_h,r,e_t) = D_{hyp}(\bm{e_h} \oplus \bm{r},\bm{e_t}) \\
   &DistH: F(e_h,r,e_t) = D_{hyp}(\bm{e_h} \circ \bm{r},\bm{e_t}) \\
   &RotH: F(e_h,r,e_t) = D_{hyp}(Rot(\bm{r})\bm{e_h},\bm{e_t}) \\
   &RefH: F(e_h,r,e_t) = D_{hyp}(Ref(\bm{r})\bm{e_h},\bm{e_t}) \\
   &D_{hyp}(\bm{x},\bm{y}) = \frac{2}{\sqrt{c}}arctanh(\sqrt{c}\|-\bm{x} \oplus \bm{y}\|)
\end{align}
where $D_{hyp}$ is the hyperbolic distance, $\oplus$ is M\"{o}bius addition operation, $\circ$ is inner product operation and $c$ is the space curvature. 

We select pre-trained low-dimensional models from this group as teachers $M_T = \{M_{T1}, M_{T1}, \dots, M_{Tm}\}$, where $m$ is the number of teachers.
Furthermore, to verify the performance of hyperbolic space in multi-teacher knowledge distillation, we prepare a corresponding Euclidean model `ModelE' for each hyperbolic model `ModelH'. 

\subsection{Junior Component}
\label{sec:3.3}
Our MulDE is a general framework, in which the Junior component can be any existing KGE model.
In this paper, we focus on the low-dimensional situation, thus selecting several effective low-dimensional models as Junior, whose initialization follows their original settings, e.g., a random normal distribution.
After the knowledge distillation training, Junior can make the link prediction faster than high-dimensional models and achieve similar precision.

Conventional knowledge distillation methods are mostly used for the classification problem.
However, the link prediction task is a learning to rank problem, which is usually learned by distinguishing the target entity with randomly negative samples. 
A straightforward solution is to learn the teachers' score distributions of the positive and negative samples.
We argue that it has at least two drawbacks. First, this teaching task is too easy for multiple teachers. Every teacher will output a result close to the hard label without a noticeable difference. Second, the negative sampling can rarely hit those indistinguishable entities, 
making 
critical knowledge owned by teachers 
hard to 
pass on to the student model.
Therefore, we design a novel distilling strategy for KGE models, including two parts of supervision.

\vspace{2mm}
\noindent\textbf{Soft Label Loss.} 
Given an e-r query, Junior evaluates all candidate entities in the entity set. Then, it selects the top-K candidates $C_{top}=\{c_1, c_2, \dots, c_K|c_i\in E\}$ with higher scores $S_{top}= \{F_J(e,r,c_i) | c_i \in C_{top}\}$. At the beginning of an iteration, this prediction sequence $\{C_{top}, S_{top}\}$ is sent to every teacher model, and the Senior component returns soft labels $L_{top}$ after integrating different teacher outputs. 
We define the soft label supervision as a Kullback-Leibler (KL) divergence, the loss function is
\begin{align}
\mathcal{L}_{Soft} &= \frac{1}{N}\sum \sigma'(L_{top}) \cdot log( \sigma'(L_{top}) / \sigma'(S_{top}))
\end{align}
where $\sigma'$ is the Softmax function, and $N$ is the number of e-r queries.

\vspace{2mm}
\noindent\textbf{Hard Label Loss.} Meanwhile, a hard label supervision based on conventional negative sampling is employed:
\begin{align}
\mathcal{L}_{Hard} =& -\frac{1}{N}\sum(L_{neg} \cdot log(\sigma(S_{neg})) \\ \nonumber
 &+ (1-L_{neg}) \cdot log(1-\sigma(S_{neg})))
\end{align}
where $\mathcal{L}_{Hard}$ is a binary cross-entropy loss, $\sigma$ is the Sigmoid function, $S_{neg}$ contains scores of positive target and sampling negative ones, and $L_{neg}$ is a one-hot label vector.

Finally, Junior loss can be formulated by 
weighted sum of $\mathcal{L}_{Soft}$ and $\mathcal{L}_{Hard}$ with a hyper-parameter $\gamma$ to balance the two parts:
\begin{align}
\mathcal{L}_J = \gamma\mathcal{L}_{Soft} + (1-\gamma)\mathcal{L}_{Hard}
\end{align}

We argue that both supervisions are necessary. The hard label supervision gives the Junior model opportunity of learning independently, and randomly negative sampling helps 
handle more general queries. Meanwhile, the novel soft label supervision corrects the top-K prediction of the Junior model, encouraging it to imitate the answer of teachers for these challenging questions.
In the experiments, we evaluate four different low-dimensional Junior models mentioned in Section \ref{sec:3.3}. 
It is worth mentioning that as a generic framework, MulDE can also be applied 
to the previous KGE models, such as DistMult \cite{DistMult} and RotatE \cite{RotatE}.

\subsection{Senior Component}
\label{sec:3.4}
As the bridge between multiple teachers and the Junior component, 
the Senior component aims to integrate different score sequences from teachers and generate soft labels $L_{top}$ suitable for the Junior component.
There are two parts of input sending from teachers to the Senior component, the top-K Junior scores  $S_{top}$ and multiple score sequences $\{S_{T1}, S_{T1},\dots, S_{Tm}\}$ ($m$ is the number of teachers) from the teacher models.
All of the above sequences have length $K$ corresponding to $K$ candidate entities in $C_{top}$.
In the Senior component, we utilize two mechanisms for knowledge integration: 

\vspace{2mm}
\noindent\textbf{Relation-specific Scaling Mechanism.} 
According to previous research \cite{RealEval2020}, 
different KGE models show performance advantages in different relations. To improve the integrated soft labels, every teacher model should contribute to the relations 
that the model is good at. 
To this end, we define a trainable scaling matrix $W_{Rel} \in \mathcal{R}^{N_r \times m}$, and assign an adaptive scaling value (ranges from 0 to 1) for each teacher sequence. The $i$-th scaled sequence ${S}'_{Ti}$ is computed as:
\begin{align}
{S}'_{Ti} = {S}_{Ti} \times \sigma(W_{Rel}[r',i])
\end{align}
where $r'$ refers to the index of the relation $r$ in an e-r query.
Note that all scores in a sequence are adjusted in proportion, while their relative ranks have no changes.
Attempts have been made to adjust each triple score individually, but the Senior loss is hard to converge and fluctuates sharply.

\vspace{2mm}
\noindent\textbf{Contrast Attention Mechanism.} 
As Junior is randomly initialized, a preliminary division of the whole entity set is needed in the early model training.
At this time, the distribution of its scores would be significantly different from that of the trained teacher. A disparate soft label would block the training process. 
Therefore, at the early period of model training, Senior is encouraged to generate soft labels having a similar distribution with Junior's scores.

Specifically, we contrast the Junior sequence with each teacher sequence and evaluate their similarity.
An attention mechanism is designed to make the integrated sequence more similar to the Junior sequence. The calculation process of soft labels $L_{top}$ is as follows:
\begin{align}
&p_i = \frac{1}{K}\sum \sigma'(S_{Ti}) \cdot log( \sigma'(S_{Ti}) / \sigma'(S_{top})) \\
&L_{top} = \sum_{i} ({S}'_{Ti} \times \sigma'(-p_i/\tau) \times m)
\end{align}
where $p_i$ is the KL divergence, a higher $p_i$ means more 
difference 
between $S_{Ti}$ and $S_{top}$.
To ensure that the contrast attention only works at the early tens of training epochs, we set a temperature parameter $\tau= exp(\lfloor n / 5 \rfloor)$. With the increase of the training epochs $n$, $\tau$ increases exponentially and $p_i$ of different teachers tend to be equal. 
So that, after the first few epochs, the soft labels $L_{top}$ are only adjusted by relation-specific scaling mechanism to achieve higher performance.

In order to train the parameters in $W_{Rel}$, 
a binary cross-entropy loss is utilized to evaluate the performance of scaled scores ${S}'_{Ti}$ in the Senior component as follows:
\begin{align}
\mathcal{L}_{S} =& -\frac{1}{N}\sum(L_{seq} \cdot log(\sigma(\sum_i {S}'_{Ti})) \\ \nonumber
 &+ (1-L_{seq}) \cdot log(1-\sigma(\sum_i {S}'_{Ti})))
\end{align}
where $L_{seq}$ is a one-hot vector in which the position of the target entity of the e-r query is 1 and the rest is 0.
Besides, if the target entity is not in the candidate set, $L_{seq}$ is the zero vector.

\subsection{Learning Algorithm}
\label{sec:3.5}
There are three roles in the proposed MulDE framework, namely {\em Teacher} (pre-trained hyperbolic models), {\em Senior Student} (integration mechanisms), and {\em Junior Student} (the target low-dimensional model).
As the teacher models have no parameter fine-tuning,
we use a combined loss function to train the two student components and minimize the loss by exploiting the Adam optimizer \cite{ADAM}.
The complete loss function $\mathcal{L}_{final}$ is as follows:
\begin{equation}
\mathcal{L}_{final} = \mathcal{L}_{S} + \mathcal{L}_{J} + \lambda \|\Theta\|_2^2
\end{equation}
where $\lambda \|\Theta\|_2^2$ is the parameter regularization term.

\begin{algorithm}
  \caption{The Learning Procedure of MulDE}
  \label{alg1}
  \begin{algorithmic}[1]
  \REQUIRE $Q$: the training KG triples; \\
  \qquad $E$:the total entity set of KG;\\
  
  \ENSURE $\Theta_J$:the parameters of Junior;\\ 
  \qquad $\Theta_S$:the parameters of Senior;\\
  // Pre-training Phase
  \FOR{i = $1,2,\cdots,m$}
  \STATE Train the $i^{th}$ teacher model $M_{Ti}$ with $Q$; \\
  \ENDFOR \\
  // Training Phase
  \STATE Initialize parameters $(\Theta_J, \Theta_S)$;\\
  \FOR{each training epoch}
  \FOR{each $(e_{in}, r, e_{miss}) \in Q$}
    \STATE $\bm{S_r}\leftarrow$ \textbf{JuniorScoreFunction}($e_{in}, r, E, \Theta_J$);\\
    \STATE $\{C_{top},S_{top}\} \leftarrow$ \textbf{GatherTopKCandidates}($\bm{S_r}, E$);\\
    \STATE $\{L_{neg},S_{neg}\} \leftarrow$ \textbf{RandomNegSamples}($\bm{S_r}, E$);\\
    \FOR{i = $1,2,\cdots,m$}
     	\STATE $S_{Ti}\leftarrow$ \textbf{TeacherScoreFunction}($M_{Ti}, e_{in}, r, C_{top}$);\\
    \ENDFOR
    \STATE $\{{S}'_{Ti}\}\leftarrow$\textbf{RelationScaling}($\{S_{Ti}\}, \Theta_S$);\\
    \STATE $L_{top}\leftarrow$\textbf{ConstrastAttention}($S_{top}, \{{S}'_{Ti}\}$);\\
  \ENDFOR
  \STATE Compute the Senior loss with $L_{top}$; \\
  \STATE Compute the Junior loss with $S_{top}$ and $S_{neg}$; \\
  \STATE Update parameters $(\Theta_J, \Theta_S)$ by gradient descent;\\
  \ENDFOR \\
  \end{algorithmic}
\end{algorithm}

The learning algorithm of MulDE is presented in Algorithm 1.
We emphasize the preciseness of this learning procedure.
It should be noted that in the pre-training and training phases, both teachers and the Junior component cannot access the triples in the validation and test sets, including the negative sampling process.

\vspace{2mm}
\noindent\textbf{Computing Efficiency.} 
We carefully consider the computing efficiency of MulDE in low-dimensional condition. 
As the parameters in teacher models are fixed and Junior is any KGE model, the only additional burden is the Senior component. Senior’s space complexity can be ignored, as the scaling mechanism only uses $N_r \times m$ parameters and the contrast attention mechanism is parameter-free. For time complexity, Senior only deals with score sequences instead of d-dimensional vectors, thus its calculation amount is less than 1/d of the time needed by Junior. 
Inevitably, the training cost of iterative distillation strategy in one epoch is more than a single KGE model, due to the calculations of teachers and the selection of topK candidates. However, we find that the convergence speed of MulDE is relatively faster, which would obviously reduce the total training times. Therefore, it is of high-efficiency to utilize MulDE to enhance existing low-dimensional KGE models. 

\begin{table}[!htb]
\caption{Statistics of the datasets.}
\centering
\label{tab:1}
\begin{tabular}{c|rrrrr}
\hline
\textbf{Dataset} & \textbf{$N_r$} & \textbf{$N_e$} & \textbf{\#Train} & \textbf{\#Valid} & \textbf{\#Test} \\
\hline
FB15k237 & $237$ & $14,541$ & $272,115$ & $17,535$ & $20,466$ \\
WN18RR & $11$ & $40,943$ & $86,845$ & $3,034$ & $3,134$ \\
\hline
\end{tabular} 
\end{table}

\begin{table*}[!tb]
\caption{Link prediction results on the two datasets. The best score of high-dimensional models \underline{underlined} and the best score of 32-dimensional models in \textbf{Bold}. The symbol `$^*$' means this score outperforms that of all previous low-dimensional models.}
\centering
\label{tab:2}
\begin{tabular}{ccc|ccc|ccc}
\hline
\multirow{2}*{\textbf{Type}} & \multirow{2}*{\textbf{Dim}} & \multirow{2}*{\textbf{Methods}} & \multicolumn{3}{c|}{\textbf{FB15K237}} & \multicolumn{3}{c}{\textbf{WN18RR}}\\
~ & ~ & ~ &  \textbf{MRR} & \textbf{Hits@10} & \textbf{Hits@1} & \textbf{MRR} & \textbf{Hits@10} & \textbf{Hits@1}\\
\hline
\multirow{10}*{High-dimensional Models} & \multirow{10}*{200d-500d} & TransE & 0.256 & 0.456 & 0.152 & 0.207 & 0.476 & 0.012 \\
~& ~& DistMult & 0.286 & 0.445 & 0.202 & 0.412 & 0.484 & 0.372 \\
~& ~& ComplEx & 0.283 & 0.447 & 0.202 & 0.431 & 0.513 & 0.395 \\
~& ~& ConvE & 0.316 & 0.501 & 0.237 & 0.430 & 0.520 & 0.400 \\
~& ~& RotatE & 0.338 & 0.533 & 0.241 & 0.476 & 0.571 & 0.428 \\
~& ~& TuckER & \underline{0.358} & 0.544 & \underline{0.266} & 0.470 & 0.526 & 0.443 \\
~& ~& QuatE & 0.348 & \underline{0.550} & 0.248 & 0.488 & 0.582 & 0.438 \\
~& ~& RefH & 0.346 & 0.536 & 0.252 & 0.461 & 0.568 & 0.404 \\
~& ~& RotH & 0.344 & 0.535 & 0.246 & \underline{0.496} & \underline{0.586} & \underline{0.449} \\
~& ~& AttH & 0.348 & 0.540 & 0.252 & 0.486 & 0.573 & 0.443 \\
\hline
\multirow{5}*{Low-dimensional Models} & \multirow{5}*{32d} & RotatE & 0.290 & 0.458 & 0.208 & 0.387 & 0.491 & 0.330 \\
~& ~& TuckER & 0.306 & 0.475 & 0.223 & 0.428 & 0.474 & 0.401 \\
~& ~& RefH & 0.312 & 0.489 & 0.224 & 0.447 & 0.518 & 0.408 \\
~& ~& RotH & 0.314 & 0.497 & 0.223 & 0.472 & 0.553 & 0.428 \\
~& ~& AttH & 0.324 & 0.501 & 0.236 & 0.466 & 0.551 & 0.419 \\
\hline 
\multirow{8}*{Our MulDE Models} & \multirow{8}*{32d} & TransH & 0.308  & 0.488  & 0.217  & 0.231  & 0.518  & 0.081 \\
~& ~ & MulDE-TransH & 0.328$^*$  & 0.511$^*$  & 0.236$^*$  & 0.267  & 0.540  & 0.094 \\
~& ~ & DistH & 0.293  & 0.474  & 0.202  & 0.439  & 0.511  & 0.399 \\
~& ~ & MulDE-DistH & 0.326$^*$  & 0.509$^*$  & 0.235  & 0.460  & 0.545  & 0.417 \\
~& ~ & RefH & 0.302  & 0.474  & 0.215  & 0.453  & 0.526  & 0.414 \\
~& ~ & MulDE-RefH & 0.325$^*$  & 0.508$^*$  & 0.233  & 0.479$^*$  & 0.569$^*$  & 0.434$^*$ \\
~& ~ & RotH & 0.310  & 0.489  & 0.221  & 0.463  & 0.547  & 0.416 \\
~& ~ & MulDE-RotH & \textbf{0.328}$^*$  & \textbf{0.515}$^*$  & \textbf{0.237}$^*$  & \textbf{0.481}$^*$  & \textbf{0.574}$^*$  & \textbf{0.433}$^*$  \\

\hline
\end{tabular}
\end{table*}

\section{Experiments}
\label{sec:4}

\subsection{Experimental Setup}

\textbf{Datasets.} Our experimental studies are conducted on two commonly used datasets.
WN18RR \cite{WNDS} is a subset of the English lexical database WordNet \cite{WordNet}.
FB15k237 \cite{Toutanova2015} is extracted from Freebase including knowledge facts about movies, actors, awards, and sports. It is created by removing inverse relations,
because many test triples can be obtained simply by inverting triples in the training set.
The statistics of the datasets are given in Table \ref{tab:1}. 
`Train', `Valid', and `Test' refer to the amount of triples in the training, validation, and test sets.

\vspace{2mm}
\noindent\textbf{Baselines.}
We implement MulDE by employing four hyperbolic KGE models as student or teacher models, including TransH, DistH, RefH, and RotH. 
Although AttH \cite{GoogleAttH} achieves better performance in one or two metrics, it is a combination of RefH and RotH.  
To verify the influence of the hyperbolic space, we also pre-train a corresponding Euclidean model `ModelE' for each hyperbolic model `ModelH'. 

The compared KGE models are in two classes:
(1) High dimensional models, including the state-of-the-art KGE methods in Euclidean space or Hyperbolic space. All of them adopt embedding vectors in more than 200 dimensions.
(2) Low-dimensional models, including RotH, RefH and AttH proposed by Chami et al. in \cite{GoogleAttH}. 
Benefiting from hyperbolic vector space, they have shown obvious advantages in the low-dimensional condition. 

\vspace{2mm}
\noindent\textbf{Implementation Details.} All experiments are performed on Intel Core i7-7700K CPU @ 4.20GHz and NVIDIA GeForce GTX1080 Ti GPU, and are implemented in Python using the PyTorch deep learning framework \cite{Pytorch}.

We select the hyper-parameters of our model via grid search according to the metrics on the validation set. 
For the teacher models, we pre-train different teachers with the embedding dimensions among $\{64, 128, 256, 512\}$. According to the preliminary analysis, we set the embedding dimensions 
as 
64 in the main experiments.
We select the learning rate among $\{0.0005, 0.001, 0.005\}$, and the number of negative samples among $\{8, 50, 255\}$.

For the MulDE framework, we empirically select the Junior's embedding dimensions among $\{8, 16, 32, 64\}$, the length $K$ of prediction sequences among $\{100, 300, 500\}$,
the learning rate among $\{0.0005, 0.001, 0.005\}$,
and the balance hyper-parameter $\gamma$ among $\{0.01, 0.1, 0.5\}$.

\vspace{2mm}
\noindent\textbf{Evaluation Metrics.}
For the link prediction experiments, we adopt two evaluation metrics:
(1) MRR (Mean Reciprocal Rank), the average inverse rank of the test triples, and
(2) Hits@N, the proportion of correct entities ranked in top N.
Higher MRR and Hits@N mean better performance.
Following the previous works, we process the output sequence in the Filter mode. It is worth mentioning that, for pre-trained models, we only remove those entities appearing in the training dataset.

\subsection{Link Prediction Task}
We first evaluate our model in the 32-dimensional vector space, which is the same as the low-dimensional setting of Chami et al. \cite{GoogleAttH}. MulDE-modelH represents the model whose Junior model is a 32-dimensional hyperbolic model, and its four teachers, TransH, DistH, RotH and RefH, are pre-trained in the 64-dimensional space. 
We compare distilled Junior models with its original results, as well as with the state-of-the-art models in both low and high dimensions. The experimental results are shown in Table \ref{tab:2}. 

From the table, we can have the following observations. The four different Junior models trained by MulDE significantly outperform their original performance on the both 
datasets. The MRR and Hits@10 of all four models have an average 5\% increase. Especially, the Hits@10 of RotH improves from 0.547 to 0.574 on WN18RR, and the Hits@1 of DistH improves from 0.202 to 0.235 on FB15k-237. The results illustrate the effectiveness of knowledge distillation for low-dimensional embedding. 

Compared with previous low-dimensional models, MulDE-RotH achieves the state-of-the-art results in all metrics on the two datasets. Although the AttH model combines both RotH and RefH, it is weaker than our next-best model MulDE-RefH in 5 metrics. Besides, the TransH model only utilizes a simple 
scoring function, but exceeds all previous low-dimensional models on the FB15k-237 dataset after the knowledge distillation training.

Compared with the high-dimensional models, the 32d MulDE-RotH exceeds multiple 200-dimensional models, including TransE, DistMult, ComplEx, and ConvE. Although its performance is lower than some of the latest RotatE, TuckER and QuatE models, MulDE-RotH has shown strong competitiveness in some metrics with much less parameters. Significantly, the Hits@10 of MulDE-RotH on WN18RR outperforms most state-of-the-art models, and its MRR and Hits@1 metrics are very close to those of the high-dimensional RotH model on the two datasets.

\begin{table}[!tb]
\caption{The improvements of MulDE-RotH on WN18RR with different teacher and student dimensions. Bold indicates better RotH results with same dimension, while {\em Growth} is the increase rate between the two models in the same region.}
\centering
\label{tab:3}
\begin{tabular}{cc|ccc}
\hline
\textbf{Methods} & \textbf{Dim} &  \textbf{MRR} & \textbf{Hits@10} &  \textbf{Hits@1} \\
\hline
Teachers & 64d & 0.487  & 0.581  & 0.440  \\
\hline
Origin & 8d & 0.210  & 0.352  & 0.140 \\
Junior & 8d & \textbf{0.399}  & \textbf{0.483}  & \textbf{0.350}  \\
{\em Growth} & ~ & 90.05\%  & 37.41\%  & 150.32\%  \\
\hline
Origin & 16d & 0.412  & 0.501  & 0.358  \\
Junior & 16d & \textbf{0.464}  & \textbf{0.547}  & \textbf{0.419}  \\
{\em Growth} & ~ & 12.60\%  & 9.07\%  & 17.22\%  \\
\hline
Origin & 32d & 0.463  & 0.547  & 0.416  \\
Junior & 32d & \textbf{0.481}  & \textbf{0.574}  & \textbf{0.433}  \\
{\em Growth} & ~ & 3.91\%  & 4.91\%  & 4.14\%  \\
\hline
Origin & 64d & 0.477  & 0.564  & 0.429  \\
Junior & 64d & \textbf{0.482} & \textbf{0.579}  & \textbf{0.430}  \\
{\em Growth} & ~ & 1.05\%  & 2.77\%  & 0.21\%  \\
\hline
\multicolumn{5}{c}{}\\
\multicolumn{5}{c}{\textbf{(a) Various Student Dimensions}}\\
\multicolumn{5}{c}{}\\
\multicolumn{5}{c}{}\\
\hline
\textbf{Methods} & \textbf{Dim} &  \textbf{MRR} & \textbf{Hits@10} &  \textbf{Hits@1} \\
\hline
Origin & 32d & 0.463  & 0.547  & 0.416  \\
\hline
Teachers & 64d & 0.487  & 0.581  & 0.440  \\
Junior & 32d & \textbf{0.481}  & \textbf{0.574}  & \textbf{0.433}  \\
{\em Growth} & ~ & -1.23\%  & -1.20\%  & -1.59\%  \\
\hline
Teachers & 128d & 0.488  & 0.582  & 0.437 \\
Junior & 32d & 0.480  & 0.572  & 0.430 \\
{\em Growth} & ~ & -1.64\%  & -1.71\%  & -1.60\% \\
\hline
Teachers & 256d & 0.483  & 0.578  & 0.434 \\
Junior & 32d & 0.473  & 0.569  & 0.424 \\
{\em Growth} & ~ & -2.07\%  & -1.60\%  & -2.30\% \\
\hline
Teachers & 512d & 0.479  & 0.584  & 0.423 \\
Junior & 32d & 0.469 & 0.574  & 0.412  \\
{\em Growth} & ~ & -2.09\%  & -1.71\%  & -2.60\% \\
\hline
\multicolumn{5}{c}{}\\
\multicolumn{5}{c}{\textbf{(b) Various Teacher Dimensions}}\\

\end{tabular}
\vspace{-3mm}
\end{table}

Overall, we can conclude that the MulDE framework can successfully improve the low-dimensional hyperbolic models. The performance of those distilled models is even comparable to some high-dimensional models.

\subsection{Comparison with Different Dimensions}
We further evaluate our MulDE framework with different dimensions.
For the student embedding dimensions, we still focus on low-dimensional models with dimensions lower than 100, and compare distilled models with the original ones.
For the teacher embedding dimensions, we select multiple dimension sizes over 64, including 128, 256, and 512. Then, we measure the performance gap between the 32-dimensional student and different high-dimensional teachers. The experimental results are shown in Table \ref{tab:3}.

The performance of different low-dimensional student models are shown in Table \ref{tab:3} (a).
The results show that the accuracy of the original models reduces when the embedding dimensions decrease. Although the distilled Junior models outperform their original models, 
the 8d and 16d models achieve relatively poor results. 
In contrast, from the `growth' metrics, MulDE contributes more improvements for lower-dimensional models. The MRR and Hits@1 of 8d RotH increase more than 90\% and 150\%, which is much more significant than that of 64d RotH. 
Furthermore, the performance of the 64d distilled model is very close to the 32-dimensional ones.
As a result, to save storage space in applications, 32d models can be used.

The results in Table \ref{tab:3} (b) support our preliminary analysis to some extent.
At first, the performance of the teacher ensemble is similar when its dimension exceeds 64.
The Hits@10 of 256d teachers is even worse than that of 64d.
It proves that when encoding the same KG, the capacity of the model 
tends to be stable when the embedding dimensions exceed the lower bound (e.g., near 64 for WN18RR).
Besides, with the increase of teacher dimensions, MRR and Hits@1 of the Junior model decrease, indicating that the teacher with over-high dimensions would be harder to transfer knowledge to a low-dimensional student. 
The experimental results motivate us to apply 64-dimensional teacher models, which achieve higher performance and save more pre-training costs.

\begin{table*}[!tb]
\caption{The results of ablation experiments on WN18RR and FB15k237. The student model of all variants is RotH. TopK refers to the iterative distilling strategy replacing random candidates as TopK candidates, while RS and CA refer to two mechanisms in the Senior component. KD512d and KD64d are two single-teacher models with only 512d and 64d RotH model as the teacher, respectively.}
\centering
\label{tab:4}
\begin{tabular}{c|ccc|ccc}
\hline
\multirow{2}*{\textbf{Methods}} & \multicolumn{3}{c|}{\textbf{FB15K237}} & \multicolumn{3}{c}{\textbf{WN18RR}}\\
~ &  \textbf{MRR} & \textbf{Hits@10} &  \textbf{Hits@1} & \textbf{MRR} & \textbf{Hits@10} & \textbf{Hits@1}\\
\hline
MulDE & \textbf{0.328}  & \textbf{0.515}  & \textbf{0.237} & \textbf{0.481}  & \textbf{0.574}  & \textbf{0.433}   \\
\hline
MulDE w/o TopK & 0.321  & 0.502  & 0.231  & 0.456  & 0.544  & 0.418  \\
MulDE w/o RS & 0.325  & 0.511  & 0.234  & 0.476  & 0.570  & 0.430  \\
MulDE w/o CA & 0.322  & 0.509  & 0.233  & 0.481  & 0.571  & 0.422  \\
MulDE w/o Senior & 0.323  & 0.501  & 0.232  & 0.459  & 0.532  & 0.419  \\
\hline
KD512d w/ TopK  & 0.322  & 0.502  & 0.232  & 0.469  & 0.564  & 0.421 \\
KD64d w/ TopK & 0.324  & 0.506  & 0.229  & 0.467  & 0.553  & 0.425  \\
KD64d & 0.321  & 0.498  & 0.230  & 0.452  & 0.540  & 0.414  \\
\hline
\end{tabular}
\end{table*}

\subsection{Ablation Studies about Distillation Strategy}

We further make a series of ablation experiments to evaluate different modules in MulDE.
There are three main improvements: 
an iterative distilling strategy (TopK), 
a relation-specific scaling mechanism (RS), and 
a contrast attention mechanism (CA). 
Therefore, we test the performance of MulDE with respect to the three modules.
Besides that, we also compare MulDE with single-teacher distillation in which the teacher is the same KGE model as the student with 64 or 512 dimensions. 
To make it harder still, we apply the iterative distilling strategy in these single-teacher models instead of random candidate labels. The experimental results are shown in Table \ref{tab:4}.

At first, for the iterative distilling strategy, it is clear that the performance of MulDE dramatically decreases when removing this module. Especially, the Hits@10 on WN18RR reduces from 0.574 to 0.544 (-5.3\%). This proves the necessity of using top-K labels when applying knowledge distillation in link prediction tasks. 
In contrast, although two mechanisms are helpful, the contributions of RS and CA are relatively small. Further improvement of the Senior component will be one of our future work.

Although single RS or CA has limited performance improvement, the existence of Senior is necessary for the interactive distillation strategy. As the results show that, when eliminating the Senior component, the Hits@10 on two datasets decreases to 0.501 and 0.532 respectively. We argue that it benefits from Senior integrating teacher scores adaptively. As both RS and CA provide a weighted operation to different teachers, eliminating Senior would make teacher scores and soft labels fixed in the training process. Only Junior can reduce the KL divergence in KD loss, and get stuck in a locally optimal solution easier. Oppositely, Senior can provide variable soft labels and adjust KL divergence in another direction.

The single-teacher knowledge distillation can improve the original RotH model, but its performance is 
worse 
than MulDE. The iterative distilling strategy also positively affects the single-teacher framework, which increases Hits@10 on WN18RR by around 2\%. A higher-dimensional teacher component (512d) 
can enhance the performance, but its training cost is unavoidable higher than MulDE.

Overall, the experimental results indicate the effectiveness of the three major modules in MulDE. Compared with the single-teacher strategy, our framework shows apparent improvements. 

\begin{figure}[!bt]
\centering
\includegraphics[width=0.43\textwidth]{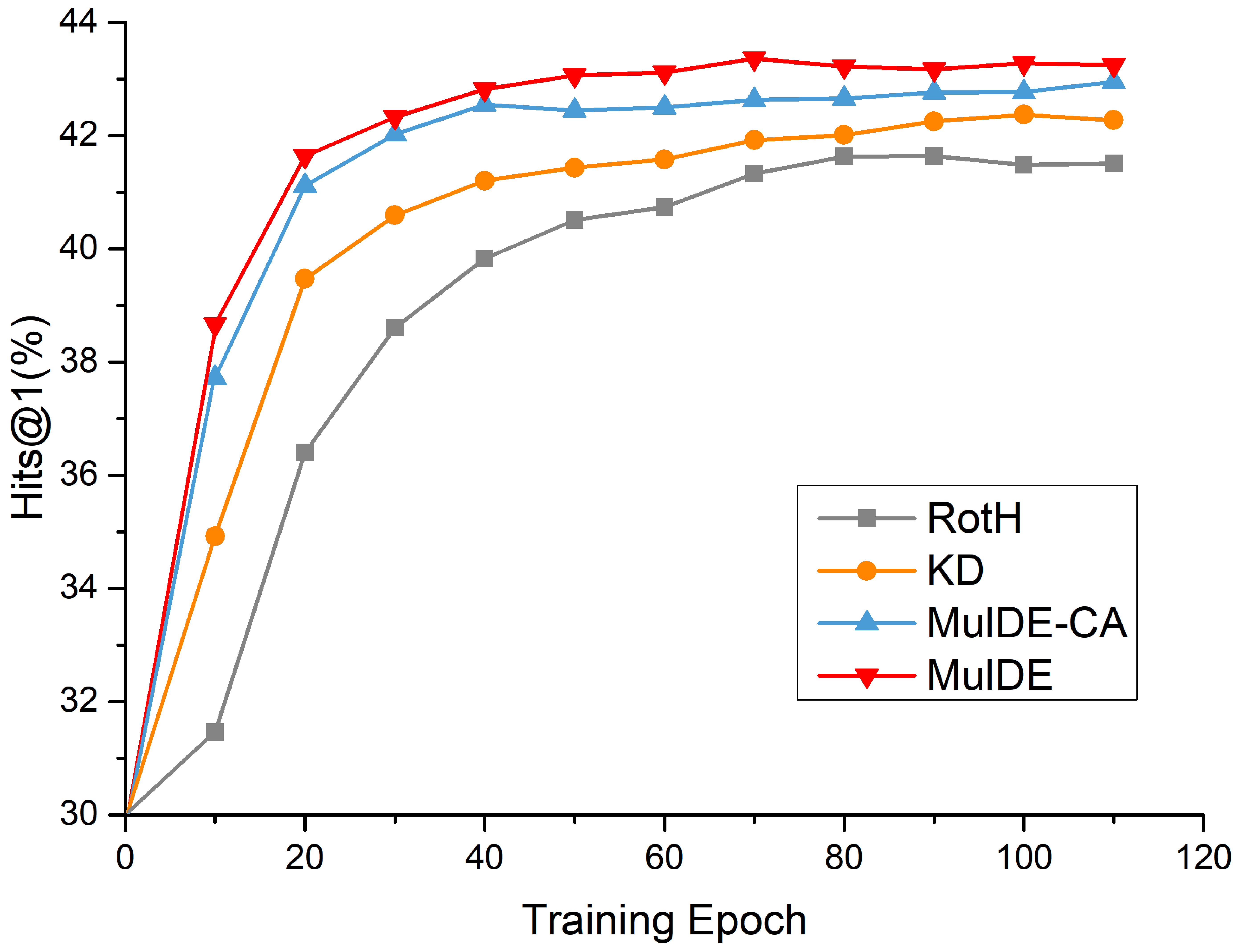}
\caption{The Hits@1 of 32-dimensional RotH as training proceeds on WN18RR. RotH is the original training mode, KD is the single-teacher knowledge distillation process, MulDE and MulDE-CA are our models with and without the contrast attention mechanism.}
\label{fig:3}
\vspace{-2mm}
\end{figure}

\section{Discussions}
\label{sec:5}
In this section, we particularly discuss several arresting questions 
around MulDE.

\vspace{2mm}
\noindent\textbf{Q1: To what extent distillation strategies accelerate student training speed?}

Fig. \ref{fig:3} shows the convergence of 32-dimensional RotH with different training processes.
As expected, the student model trains faster under the supervision of a teacher. 
We can observe that MulDE converges faster than original RotH, and achieves higher accuracy. In contrast, the single-teacher KD model is much slower and continuously increases before 100 epochs.
We also analyze the convergence of MulDE without the contrast attention mechanism, i.e., MulDE-CA.
Benefiting from contrast attention, the Hits@1 of MulDE in the early period (epoch=10) is around 3\% higher than MulDE-CA, while it also leads to a higher final accuracy from 0.422 to 0.433. This indicates that the contrast attention can reduce the gap between student scores and teacher scores in the early period.

\vspace{2mm}
\noindent\textbf{Q2: Whether the hyperbolic space contributes to the result?}

\begin{figure}[!bt]
\centering
\includegraphics[width=0.4\textwidth]{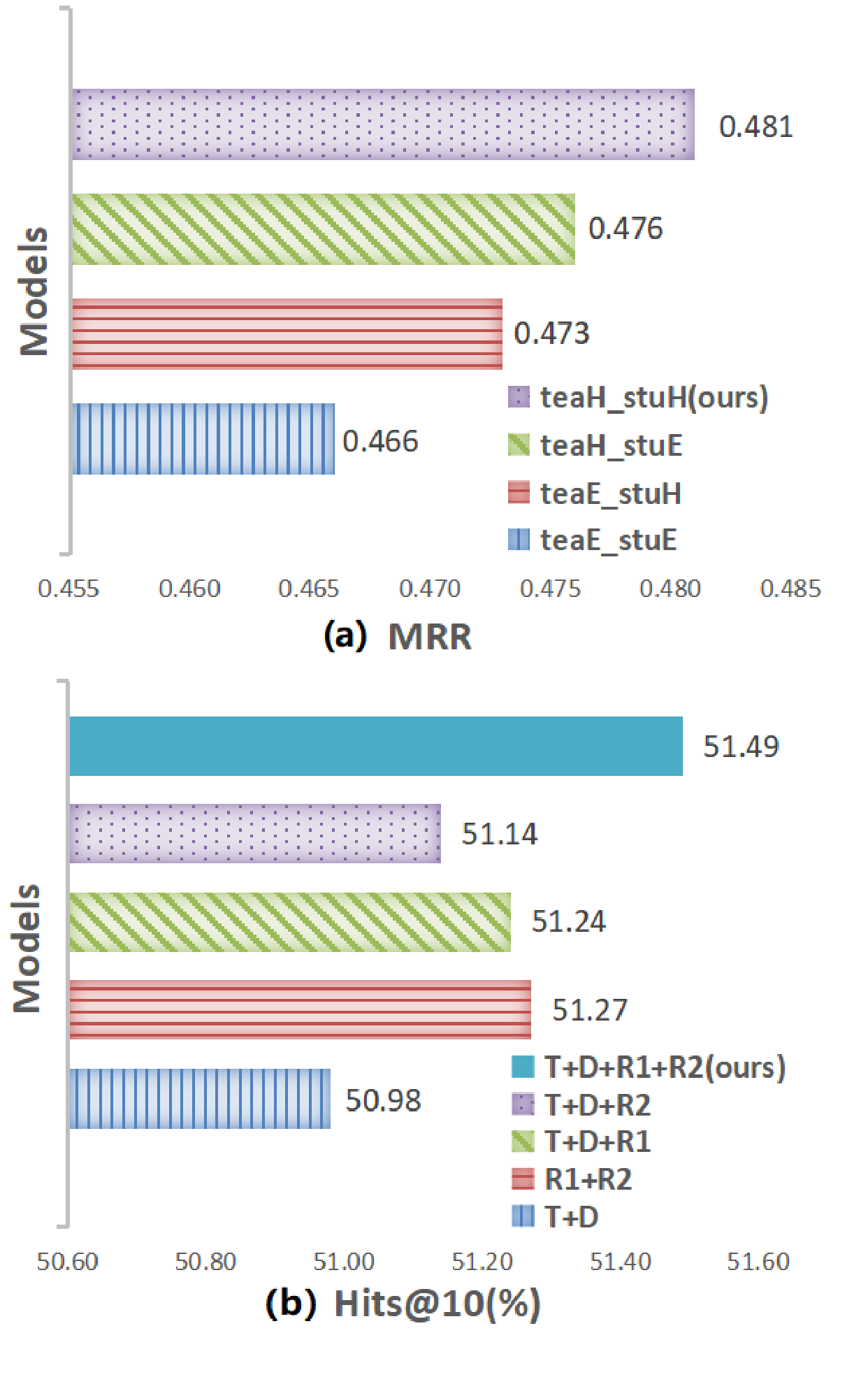}
\caption{Performance of 32-dimensional student with different variants. (a) The MRR on WN18RR with different settings of vector space. (b) The Hits@10 on FB15k237 with different teacher combinations. 
We use T, D, R1, R2 refers to TransH, DistH, RotH and RefH.}
\label{fig:4}
\vspace{-2mm}
\end{figure}

We evaluate the MulDE framework by employing hyperbolic KGE models based on their performance in low dimensions. 
To answer the question, we analyze the contributions of hyperbolic models when they are teachers or students, and also verify the effectiveness of MulDE in Euclidean space. The results are shown in Fig. \ref{fig:4} (a). Here `teaH' and `stuH' represent using hyperbolic models, while `teaE' and `stuE' 
indicate using corresponding Euclidean models. 
The teaH\_stuH model is equal to MulDE-RotH, and the other variants assign different space types to teachers and students.
Among the four models, teaE\_stuE is weaker than the other three, which indicates the advantage of hyperbolic space. Even so, the 0.466 Hits@1 is still equal to that of the previous AttH.
Considering the middle two models, we can conjecture that using hyperbolic space in the teacher components is more effective than using a hyperbolic student model.

\vspace{2mm}
\noindent\textbf{Q3: How about the contribution of different teacher choices on results?}

The complete MulDE framework utilizes all four hyperbolic models as teachers, because the whole ensemble can obtain better accuracy than the other combinations. We further analyze the contribution of every single teacher model in the ensemble. 
As shown in Fig. \ref{fig:4} (b), the performance of MulDE outperforms the other four teacher combinations. 
In terms of combining two teachers, only using TransH and DistH is obviously poorer than using RotH and RefH, which indicates the importance of the latter models. Comparing two models using three teachers, we find that RotH contributes more than RefH. It is reasonable because the accuracy of the original RotH is already higher than RefH.
We also trial other combinations of the four models (e.g., T+R1, D+R1, T+R2, D+R2), and the experimental results prove the contribution of every teacher model in MulDE. 
One of our future work will be discovering a better teacher combination by adding other different models.

\begin{table*}[!tb]
\caption{The prediction results of different models about two triples. The correct candidate is in bold.}
\centering
\small
\label{tab:5}
\begin{tabular}{cp{2.5cm}|l}
\hline
\textbf{Query} & \textbf{Methods} & \textbf{Top 5 Candidates}\\
\hline
\multirow{3}*{\tabincell{l}{(The X-Files,\\/tv/tv\_program/genre)}} & \textbf{RotH} &  Psychological Thriller, Fantasy, Adventure Film, Comedy Drama, Period Piece\\
& \textbf{MulDE-RotH} & Fantasy, \textbf{Crime Fiction}, Horror, Adventure Film, Comedy Drama\\
& \textbf{Teachers} & Fantasy, \textbf{Crime Fiction}, Psychological Thriller, Horror, Adventure Film\\
\hline
\multirow{3}*{\tabincell{l}{(Comcast, \\/business/business\\\_operation/industry)}} & \textbf{RotH} &  Software, Computer hardware, Retail, Manufacturing, \textbf{Telecommunications}\\
& \textbf{MulDE-RotH} & \textbf{Telecommunications}, Retail, Software, Video Game, Computer Hardware\\
& \textbf{Teachers} & \textbf{Telecommunications}, Software, Computer Hardware, Manufacturing, Retail\\
\hline
\end{tabular}
\end{table*}

\vspace{2mm}
\noindent\textbf{Q4: Which parts of prediction results obtain improvement after enhancing by MulDE?}

The failed prediction of KGE models may be caused by various reasons, e.g., KG in-completion, similar entities, and data sparseness. As MulDE can effectively improve Junior's accuracy, we deeply explore those improved prediction samples in evaluating phase.
Tabel \ref{tab:5} shows two groups of prediction results about two triples on the FB15k237 test set.

In the first example, when predicting the genre of this science-fiction show, the original RotH model fails to rank the target `Crime Fiction' in the top five candidates, while both the teacher ensemble and our MulDE-RotH rank the target at second. 
It indicates that MulDE-RotH outperforms RotH, benefiting from the supervision of teachers.
Although MulDE-RotH still predicts a wrong tail entity, `Fantasy' is reasonable in the human view.
A similar situation is shown in the second example, in which MulDE-RotH successfully finds the correct answer `Telecommunications' after training by MulDE. It should be noted that MulDE-RotH does not directly reproduce teachers, as its results are different from teachers'.

In case studies on the FB15237 test set, the teacher ensemble correctly predicts 1,907 more samples than 32-dimensional original RotH, while the number of improved samples for the 32-dimensional MulDE-RotH is 1,808 and 75.6\% of samples are same with teachers'. The results indicate that a staggering proportion of improved samples of MulDE-RotH are benefited from the supervision of teachers.

\section{Related Work}
\label{sec:6}
In this section, we discuss recent research 
advance in the KGE domain, and introduce relevant Knowledge Distillation research.

\subsection{Knowledge Graph Embeddings}
Predicting new triples using KGE methods has been an active research topic over the past few years \cite{OlddogICLR20}.
Benefiting from the simple calculation and good performance, some early methods such as TransE \cite{TransE}, DistMult \cite{DistMult}, ComplEx \cite{ComplEx} have been widely used in various AI tasks. With the rise of deep learning, several CNN-based methods have been proposed, such as ConvE \cite{ConvE} and ConvKB \cite{ConvKB}. These methods 
achieve good performance in link prediction tasks, but heavily depend on more parameters \cite{CompGCN, RGCN}.
Recently, there are several non-neural methods proposed.
RotatE \cite{RotatE}, inspired by Euler's identity, can infer various relation patterns with a new rotation-based scoring function. 
Balazevic et al. \cite{TuckER} propose a linear model based on Tucker decomposition of the binary tensor representation of knowledge graph triples.
QuatE \cite{QuatE} and OTE \cite{OTE} further improve the RotatE method. The former introduces more expressive hypercomplex-valued representations in the quaternion space, while the latter (i.e., OTE) extends the RotatE method from a 2-D complex domain to a high dimensional space with orthogonal transforms to model relations.

Since a single KGE model faces a contradiction between complexity and performance, a straightforward solution is to take an ensemble of multiple KGE methods to predict the same target. Krompass et al. \cite{Ensemble1} designed a simple ensemble consisting of multiple KGE methods, in which the final prediction results rely on an average score of all sub-modules.
Wang et al. \cite{Ensemble2} proposed a relation-level ensemble that combines multiple individual models to pick the best model for each relation.

\subsection{KGE Model Compression}
Research related to the KGE model compression is relatively recent and new, 
featuring 
two representative approaches.
One possible solution is to compress pre-trained high-dimensional models. Sachan \cite{KGCompress} utilized Embedding Compression methods to convert high-dimensional continuous vectors into discrete codes. Although generating a compressed model retaining high-accuracy knowledge, a pre-trained high-dimensional model is necessary. Furthermore, the compressed model with discrete vectors cannot continue the training task when the KG is modified.
Another solution 
introduces new theories to improve low-dimensional KGE models directly. For example, Chami et al. \cite{GoogleAttH} introduced the hyperbolic embedding space with trainable curvature, and proposed a class of hyperbolic KGE models, which outperforms previous Euclidean-based methods in low-dimension. 
However, the limited number of parameters inevitably declines the model performance, and the 
low-dimensional 
models cannot utilize the high-accuracy knowledge from high-dimensional ones.

\subsection{Knowledge Distillation}
Knowledge Distillation (KD) aims to transfer knowledge from one machine learning model (i.e., the teacher) to another model (i.e., the student) \cite{KDICLR20, KDCVPR19}.
Hinton et al. \cite{1stKD} introduced the first KD framework, which applies the classification probabilities of a trained model as soft labels and defines a parameter called ``Temperature'' ($T$) to control the soft degree of those labels.
Specifically, 
when $T \rightarrow 0$, the soft labels become one-hot vectors, i.e., the hard labels.
With the increase of $T$, the labels become softer.
When $T \rightarrow \infty$, all classes share the same probability \cite{KDWWW17, KDICCV19}. 
Inspired by this, several KD-based approaches are proposed in different research domains.
Furlanello et al. \cite{2018KD} proposed Born-Again Networks in which the student is parameterized identically to their teachers.
Yang et al. \cite{2019KD} added an additional loss term to facilitate a few secondary classes to emerge and complement to the primary class.
Li et al. \cite{2019IJCAIKD} proposed a method to distill human knowledge from a teacher model to enhance a pedestrian attribute recognition task.
To the best of our knowledge, our work is the first to apply the KD technologies in the link prediction of KGs.

\section{Conclusion}
\label{sec:7}
Recent knowledge graph embedding (KGE) models tend to apply high-dimensional embedding vectors to improve their performance.
These models 
can be hardly applied in practical knowledge graphs (KGs) due to large training costs and storage space.
In this paper, we propose MulDE, a novel multi-teacher knowledge distillation framework for knowledge graph embeddings.
By integrating multiple hyperbolic KGE models as teachers, we present a novel iterative distillation strategy to extract high-accuracy knowledge and train low-dimensional students adaptively.
The experimental results show that the low-dimensional models distilled by MulDE outperform the state-of-the-art models on two widely used datasets.
Compared with general single-teacher knowledge distillation methods, MulDE can accelerate student training speed.

These positive results encourage us to explore the following further research activities in the future:
\begin{itemize}
\item We will further research the relationship between embedding dimensions and the KG scale, and explore the influence of different relation transformations.

\item To take advantages of multiple teachers, we will further improve the knowledge integration in the Senior component, and achieve higher-accuracy soft labels in the low dimensional models.

\item Regarding the choice of multiple teachers, we will 
further analyze the effectiveness of the ensemble. In particular, we will 
investigate new teacher combinations by considering emerging relation transformations.
\end{itemize}

\begin{acks}
This research is supported by the National Natural Science Foundation in China
(Grant: 61672128) and the Fundamental Research Fund for Central University (Grant: DUT20TD107).
Quan Z. Sheng has been partially supported by Australian Research Council (ARC) Future Fellowship Grant FT140101247, and Discovery Project Grant DP200102298.
\end{acks}

\bibliographystyle{ACM-Reference-Format}
\bibliography{bibtex}

\end{document}